\documentclass[11pt,letterpaper]{article}
\usepackage{emnlp2017}
\usepackage{times}
\usepackage{graphicx,multirow}
\usepackage{colortbl}
\usepackage{url}
\usepackage{color}
\usepackage{float}
\usepackage{booktabs}
\usepackage{tabularx}
\usepackage{slashbox}

\usepackage{subfig}
\usepackage{multirow}
\usepackage{latexsym}
\usepackage{amsmath}
\usepackage{amssymb}
\usepackage{pgfplots}
\usepackage{algorithm}
\usepackage{algorithmic}
\DeclareMathOperator*{\argmax}{arg\,max}

\newcommand{\tabincell}[2]{\begin{tabular}{@{}#1@{}}#2\end{tabular}}

\def\bz{\mathbf{z}}
\def\h{\mathbf{h}}
\def\e{\mathbf{e}}

\def\bb{\mathbf{b}}
\def\i{\mathbf{i}}
\def\o{\mathbf{o}}
\def\f{\mathbf{f}}

\def\bc{\mathbf{c}}

\def\bW{\mathbf{W}}

\def\bU{\mathbf{U}}
\def\bA{\mathbf{A}}

\allowdisplaybreaks

\newenvironment{itemize*}%
  {\begin{itemize}%
    \setlength{\itemsep}{0pt}%
    \setlength{\parskip}{0pt}}%
  {\end{itemize}}
  \newenvironment{enumerate*}%
  {\begin{enumerate}%
    \setlength{\itemsep}{0pt}%
    \setlength{\parskip}{0pt}}%
  {\end{enumerate}}

\emnlpfinalcopy
\def\emnlppaperid{1423}

\title{DAG-based Long Short-Term Memory for Neural Word Segmentation}
\author{Xinchi Chen, Zhan Shi, Xipeng Qiu\thanks{{ }{ }Corresponding author.}, Xuanjing Huang\\
 Shanghai Key Laboratory of Intelligent Information Processing, Fudan University\\
School of Computer Science, Fudan University\\
825 Zhangheng Road, Shanghai, China\\
\{xinchichen13,zshi16,xpqiu,xjhuang\}@fudan.edu.cn}

\date{}

\begin{document}

\maketitle

\begin{abstract}
Neural word segmentation has attracted more and more research interests for its ability to alleviate the effort of feature engineering and utilize the external resource by the pre-trained character or word embeddings. In this paper, we propose a new neural model to incorporate the word-level information for Chinese word segmentation. Unlike the previous word-based models, our model still adopts the framework of character-based sequence labeling, which has advantages on both effectiveness and efficiency at the inference stage. To utilize the word-level information, we also propose a new long short-term memory (LSTM) architecture over directed acyclic graph (DAG). Experimental results demonstrate that our model leads to better performances than the baseline models.
\end{abstract}

\section{Introduction}

Chinese word segmentation (CWS) is a preliminary and important task for Chinese natural language processing (NLP). Currently, the state-of-the-art methods are based on statistical supervised learning algorithms, which can be further divided into character-based \cite{Xue:2003,Peng:2004,Zhao:2006} and word-based \cite{Andrew:2006,zhang2007chinese,sun2009discriminative} methods. The character-based methods regard word
segmentation as a sequence labeling problem, and each character is assigned a segmentation tag to indicate its relative position inside word. It is rather difficult to utilize the word-level features in the character-based methods. Instead, the word-based methods directly score the entire candidate segmented word sequence, which can fully utilize both the character-level and word-level information \cite{sun2010word}.

Recently, there are several neural models applied to CWS task for their ability to minimize the effort in feature engineering. These models are still divided into character-based \cite{zheng2013deep,pei2014maxmargin,chen2015long,ma2015accurate,xu2016dependency,yao2016bi} and word-based \cite{cai2016neural,zhang2016transition,liu2016exploring} methods. Table \ref{tb:overview} gives an overview of some representative word segmentation methods.

\begin{table}
  \centering \footnotesize    \setlength{\tabcolsep}{2pt}
    \begin{tabular}{l|c|c}
     \hline
      {\tiny\backslashbox{Feature}{Model}}& \tabincell{l}{Character-based\\sequence labeling} & Other\\
     \hline
     \tabincell{l}{Character\\} & \tabincell{l}{\citet{Xue:2003}\\\citet{Zhao:2006}\\\citet{zheng2013deep}$^*$\\\citet{chen2015long}$^*$\\} & \citet{ma2015accurate}$^*$ \\
     \hline
     Word & Ours$^*$ & \tabincell{l}{\citet{Andrew:2006}\\\citet{zhang2007chinese}$^*$\\     \citet{zhang2016transition}$^*$\\\citet{cai2016neural}$^*$\\\citet{liu2016exploring}$^*$}\\
     \hline
   \end{tabular}
   \caption{Overview of the word segmentation methods. $^*$~indicates that the model is neural-based. }\label{tb:overview}
\end{table}

Although the word-based information are effective in CWS, it is nontrivial to incorporate these information into the character-based sequence labeling. Existing word-based CWS methods adopt different inference methods, such as transition-based methods \cite{zhang2007chinese,zhang2016transition}, Semi-Markov conditional random field (semi-CRF) \cite{Andrew:2006,liu2016exploring}, or discriminative structured learning \cite{cai2016neural}. Among these methods, the number of candidate segmentations grows exponentially with the sequence length. Therefore, beam-search is often used to reduce error propagation. Besides, the maximum length of words is also constrained (usually less than 5) to reduce the time complexity. These two strategies usually result in an inexact inference.


In this paper, we propose a neural-based architecture for Chinese word segmentation, which integrates the word-level information to the framework of character-based sequence labeling. Specifically, given a character sequence, we build a directed acyclic graph (DAG) using a vocabulary. Each edge in the DAG denotes that its covering subsequence is a word in the vocabulary. Then, we propose a DAG-structured long short-term memory (DAG-LSTM), and the input of each position consists of the embeddings of the character and words. By using DAG-LSTM, we can model the contextual information for each position based on both the character-level and word-level information. Extensive experiments on three popular CWS datasets show that our architecture achieves better performance with the original LSTM model.

The contributions of this paper could be summarized as follows.
\begin{itemize*}
  \item The proposed DAG-LSTM can effectively integrate the word-level information for character-based CWS.
  \item We propose a dropout strategy for in-vocabulary (IV) words, so that our model could deal with the out-of-vocabulary (OOV) words. Thus, our model can easily incorporate an external vocabulary to boost the performance even without the embeddings of the OOV words.
\end{itemize*}

\section{Long Short-Term Memory Networks for Chinese Word Segmentation}
Chinese word segmentation task is usually regarded as a character based sequence labeling problem \cite{zheng2013deep,pei2014maxmargin,ma2015accurate,xu2016dependency,yao2016bi,cai2016neural,zhang2016transition}. Specifically, each character in a sentence is labeled as one of $\mathcal{T} =  \{B, M, E, S\}$, indicating the begin, middle, end of a word, or a word with single character. In this section, we will introduce the conventional long short-term memory networks \cite{chen2015long} for character based Chinese word segmentation task.

\begin{figure}[t]
  \centering
  \includegraphics[width=0.45\textwidth]{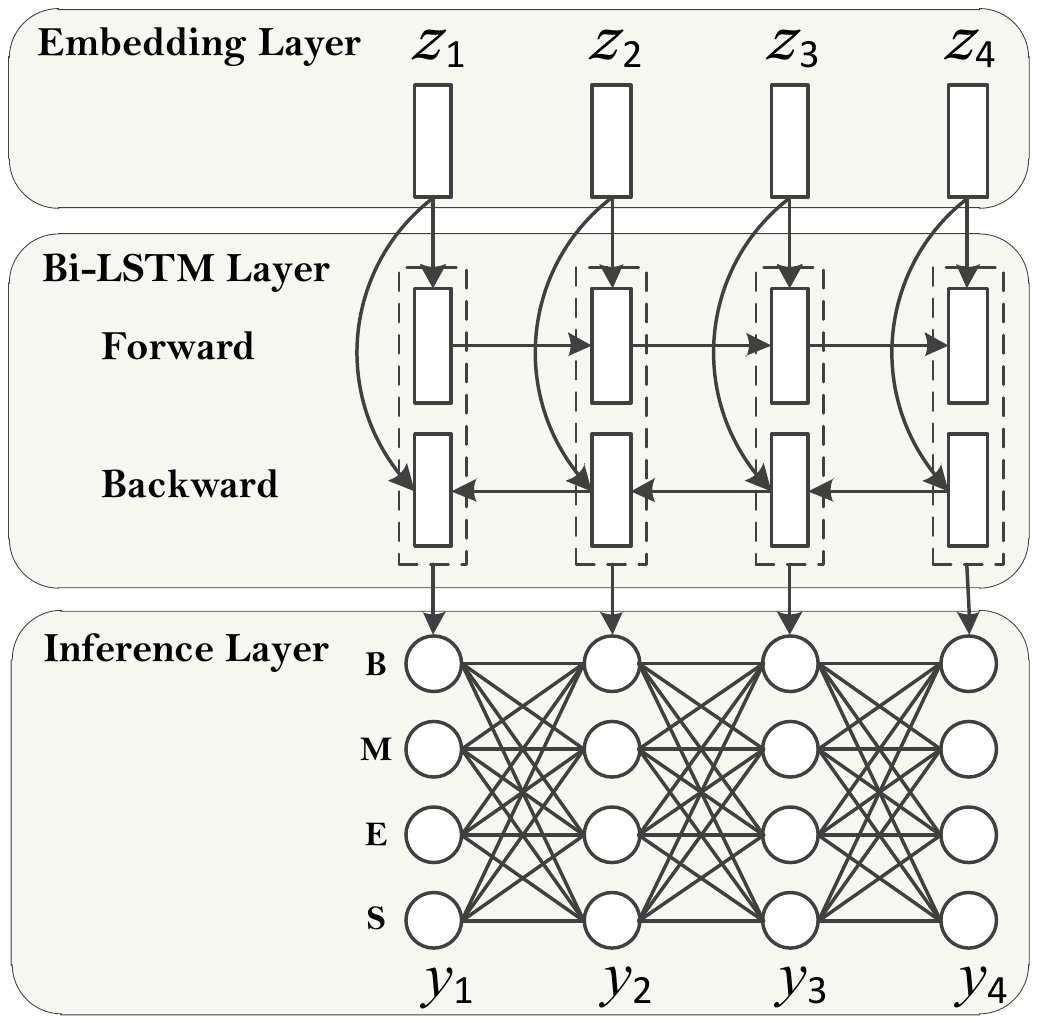}
  \caption{General neural architecture for character based Chinese word segmentation task.  }\label{fig:rnn_cws}
\end{figure}

Figure \ref{fig:rnn_cws} gives a general architecture of the neural CWS, which could be characterized by three components: (1) an embedding layer; (2) a LSTM layer and (3) an inference layer.

\begin{figure}[t]
  \centering
  \subfloat[Uni-gram]{
  \includegraphics[width=0.20\textwidth]{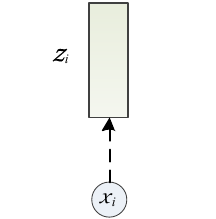} \label{fig:unigram}
  }
  \hspace{1em}
  \subfloat[Bi-gram]{
  \includegraphics[width=0.20\textwidth]{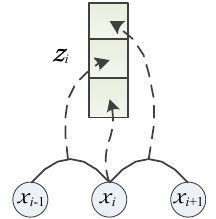} \label{fig:bigram}
  }
  \caption{Uni-gram and bi-gram embeddings.}
\end{figure}
\subsection{Embedding Layer}
In the neural models, the first step is to map discrete language symbols to distributed inputs, which is usually a concatenation operation on embeddings in different granularities. Specifically, given a sequence with $n$ characters $X = \{x_1, \cdots, x_n\}$, the distributed input $\bz_i$ for each position $i$ is usually assigned with a uni-gram embedding or a bi-gram embedding.

\paragraph{Uni-gram Embedding} \label{sec:unigram}
Uni-gram embedding input $\bz_i$ only uses the embedding of the character $x_i$ as shown in Figure \ref{fig:unigram}. Specifically, $\bz_i$ could be expressed as:
\begin{equation}
  \bz_i = \e_{x_i},
\end{equation}
where $\e_{x_i} \in \mathbb{R}^{d_e}$ is derived by a looking up operation in a embedding matrix $\mathbf{E} \in \mathbb{R}^{|\mathcal{V}_{\text{train}}| \times d_e}$. $d_e$ is a hyper-parameter indicating the size of the embedding. $\mathcal{V}_{\text{train}}$ is the vocabulary set of the train set.
\paragraph{Bi-gram Embedding} \label{sec:bigram}
Besides the uni-gram character embeddings, the bi-gram character embeddings are often used. As previous work \cite{pei2014maxmargin,chen2015long} reports, the bi-gram embeddings can significantly boost the performance of CWS. The bi-gram embedding input of each $\bz_i$ additionally considers the embedding of the bi-gram compositions at $i$-th position as shown in Figure \ref{fig:bigram}. Specifically, $\bz_i$ could be expressed as:
\begin{equation}
  \bz_i = \e_{x_i} \oplus \e_{(x_{i-1},x_i)} \oplus \e_{(x_{i},x_{i+1})},
\end{equation}
where $\oplus$ is a concatenation operation. $(x_{i-1},x_i)$ indicates the bi-gram unit composed of two consecutive characters $x_{i-1}$ and $x_i$. Notably, if $(x_{i-1},x_i)$ is not in $\mathcal{V}_{\text{train}}$, we will use a special symbol ``<OOV>'' (indicating a out-of-vocabulary term) instead.

\subsection{LSTM Layer}
Long short-term memory network (LSTM) \cite{hochreiter1997long} is a typical type of recurrent neural network (RNN) \cite{Elman:1990}, and specifically addresses the issue of learning long-term dependencies and gradient vanishing problem. Specifically, LSTM, with input gate $\i$, output gate $\o$, forget gate $\f$ and memory cell $\bc$, could be expressed as:
\begin{align}
\mathbf{i}_i &=   \sigma({\bW}^{(\mathbf{i})} \bz_i + \bU^{(\mathbf{i})} {\mathbf{h}}_{i-1} + \bb^{(\mathbf{i})}),\\
\mathbf{o}_i &=   \sigma({\bW}^{(\mathbf{o})} \bz_i + \bU^{(\mathbf{o})} {\mathbf{h}}_{i-1} + \bb^{(\mathbf{o})}),\\
\mathbf{f}_i &=   \sigma({\bW}^{(\mathbf{f})} \bz_i + \bU^{(\mathbf{f})} {\mathbf{h}}_{i-1} + \bb^{(\mathbf{f})}),\\
\tilde{\mathbf{c}}_i &=   \phi({\bW}^{(\tilde{\mathbf{c}})} \bz_i + \bU^{(\tilde{\mathbf{c}})} {\mathbf{h}}_{i-1} + \bb^{(\tilde{\mathbf{c}})}),\\
\mathbf{c}_i    &= {\tilde{\mathbf{c}}}_i \odot \mathbf{i}_i + \mathbf{c}_{i - 1} \odot \mathbf{f}_i, \\
{\mathbf{h}}_i &= \mathbf{o}_i \odot \phi( \mathbf{c}_i ),
\end{align}
where $\bW_g \in \mathbb{R}^{(d_e + d_h) \times 4d_h}$ and $\bb_g \in \mathbb{R}^{4d_h}$ are trainable parameters. $d_h$ is a hyper-parameter, indicating the hidden state size. Function $\sigma(\cdot)$ and $\phi(\cdot)$ are sigmoid and tanh functions respectively.
\paragraph{Bi-LSTM}
In order to incorporate information from both sides of sequence, we use bi-directional LSTM (Bi-LSTM) with forward and backward directions. Specifically, each Bi-LSTM unit can be expressed as:
\begin{equation}
\h_i = \overrightarrow{\h}_i \oplus {\overleftarrow{\h}_i}
\end{equation}
where $\overrightarrow{\h}_i$ and $\overleftarrow{\h}_i$ are the hidden states at $i$-th position of the forward and backward LSTMs respectively. $\oplus$ is the concatenation operation.

\subsection{Inference Layer}
In inference layer, we introduced the transition score $\bA_{ij}$ for measuring the possibility of jumping from tag $i \in \mathcal{T}$ to tag $j \in \mathcal{T}$ \cite{collobert2011natural,zheng2013deep,pei2014maxmargin}. The sentence-level prediction score is the sum of the tag transition score and the network tagging score.

Specifically, given a input sentence $X = \{x_1, \cdots, x_n\}$ with a predicted tag sequence $Y = \{y_1, \dots, y_n\}$, the sentence-level prediction score $s(X, Y; {\Theta})$ could be calculated as:
\begin{equation}
    s(X, Y; {\Theta}) = \sum_{i = 1}^n \left(\bA_{y_{i-1}y_{i}}+f(y_{i})\right), \label{eq:decode}
\end{equation}
where ${\Theta}$ indicates all trainable parameters of our model. $f(y_{t})$ is the network tagging score, which could be formalized as:
\begin{equation}
  f(y_{i}) = \bW_s \h_i + \bb_s,
\end{equation}
where $\bW_s \in \mathbb{R}^{|\mathcal{T}| \times d_h}$ and $\bb_s \in \mathbb{R}^{|\mathcal{T}|}$.
\begin{figure*}[t]
  \centering
  \includegraphics[width=0.95\textwidth]{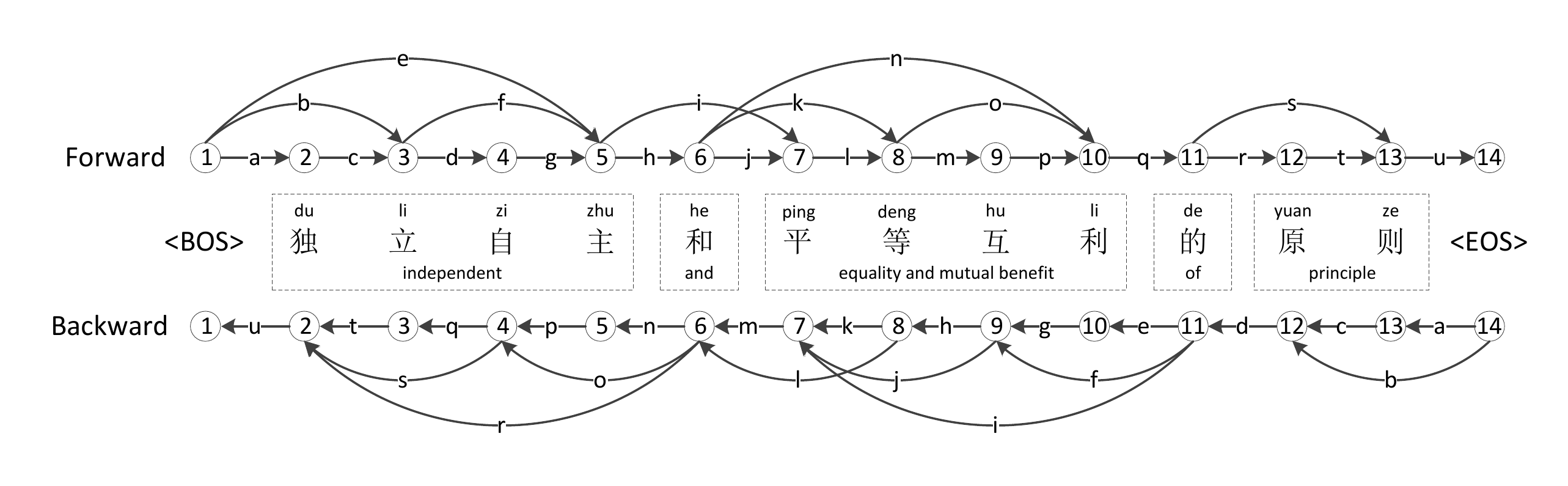}
  \caption{Building DAG for a sequence. The nodes in the DAG are indexed from 1 to 14, and the edges in the DAG are indexed from a to u. Each edge is associated with a word in train set vocabulary $\mathcal{V}_{\text{train}}$. For instance, the edge $e$ in the forward DAG is associated with the word `` 独立自主(independent)'', while the edge $f$ in the backward DAG is associated with the word ``互利 (mutual benefit)''. Here, ``<BOS>'' and ``<EOS>'' are two special symbols indicating the begin and the end of a sentence respectively.}\label{fig:dag}
\end{figure*}
\section{Building DAG for Sequence}
DAG is short for directed acyclic graph. Actually,  in the preprocess phase, we could build a DAG for every sequence in the corpus (including the train set, development set and test set) based on the vocabulary $\mathcal{V}_{\text{train}}$ of train set.  Figure \ref{fig:dag} gives an illustration of the DAG of a sequence. Each edge over nodes in the graph indicates that there is a word in $\mathcal{V}_{\text{train}}$. Notably, the forward and backward DAGs are asymmetric in our model as shown in Figure \ref{fig:dag}.

In order to build DAG more efficiently, we adopt Aho Corasick algorithm \cite{aho1975efficient} for fast DAG construction. Firstly, an automata is built using the given vocabulary $\mathcal{V}_{\text{train}}$. Thus, a DAG can be built for a character sequence by the built automata in $O(n)$.

\section{DAG-LSTM for Chinese Word Segmentation}
In order to integrate the information from the word level and character level simultaneously, we proposed two types of DAG-structured long short-term memory (DAG-LSTM) neural networks upon the built DAG for CWS task: the weight sharing model and the weight independent model. The difference between them is that whether the associated weight matrix is shared when various inputs enter to the DAG-LSTM. These two models are extensions of the tree-structured LSTMs \cite{tai2015improved}. It is worth noting that \citet{zhu2016dag} also proposed a DAG-LSTM model, which adopts a different binarized merging operation.
%
%
\subsection{Model-I: Weight Sharing DAG-LSTM} \label{sec:model-I}
Given the DAG of each sequence as shown in Figure \ref{fig:dag}, we build DAG-LSTM for CWS task upon the DAG. Since the algorithm of forward DAG-LSTM is the same with the backward one, we only describe the forward one here.

In DAG-LSTM, the information flow of LSTM separate and merge along with the structure of the DAG of the sequence. Specifically, for each position $i$, we will take all words ending with the $i$-th character $x_i$ into account. Thus, we will derive a word set $\{x_{i-l+1:i}\}_{l \in L_i}$ for $i$-th position, where $L_i$ is a length set. For simplicity, we define $\bz_{i,l} = \e_{x_{i-l+1:i}} \in \mathbb{R}^{d_e}$. All word lengths $l\in L_i$ should make sure that $x_{i-l+1:i} \in \mathcal{V}_{\text{train}}$. Notably, no ``<OOV>'' terms will occur in DAG-LSTMs. Table \ref{tab:DAG-LSTM} shows the details of data flow of the forward DAG-LSTM.

\begin{table}[t] \small
\centering
\begin{tabular}{|c|c|c|c|c|}
\hline
Target & $L$&Edge ID&PreState&Input\\
\hline
${\mathbf{h}}_{1}$&$\phi$&-&-&-\\
\hline
${\mathbf{h}}_{2}$&\{1\}&a&${\mathbf{h}}_{1}$&$\e_{x_2}$\\
\hline
\rotatebox{90}{$\dots$}&\rotatebox{90}{$\dots$}&\rotatebox{90}{$\dots$}&\rotatebox{90}{$\dots$}&\rotatebox{90}{$\dots$}\\
\hline
\multirow{3}*{${\mathbf{h}}_{5}$}&\multirow{3}*{\{1,2,3\}}&e&${\mathbf{h}}_{1}$&$\e_{x_{2:5}}$\\
&&f&${\mathbf{h}}_{3}$&$\e_{x_{4:5}}$\\
&&g&${\mathbf{h}}_{4}$&$\e_{x_{5}}$\\
\hline
\rotatebox{90}{$\dots$}&\rotatebox{90}{$\dots$}&\rotatebox{90}{$\dots$}&\rotatebox{90}{$\dots$}&\rotatebox{90}{$\dots$}\\
\hline
\multirow{2}*{${\mathbf{h}}_{13}$}&\multirow{2}*{\{1,2\}}&s&${\mathbf{h}}_{11}$&$\e_{x_{12:13}}$\\
&&t&${\mathbf{h}}_{12}$&$\e_{x_{13}}$\\
\hline
${\mathbf{h}}_{14}$&\{1\}&u&${\mathbf{h}}_{13}$&$\e_{x_{14}}$\\
\hline
\end{tabular}
\caption{Information flow of the forward DAG-LSTM for the given example in Figure \ref{fig:dag}. Here, the initial state ${\mathbf{h}}_{1} = \mathbf{0}$.}\label{tab:DAG-LSTM}
\end{table}

Formally, the the weight sharing DAG-LSTMs (WS-DAG-LSTMs) could be further derived as:
\begin{align}
\tilde{\mathbf{z}}_i &= \sum_{l\in L_i}  \bz_{i,l} , \quad
\tilde{\mathbf{h}}_i = \sum_{l\in L_i}  {\mathbf{h}}_{i-l} ,\\
\mathbf{i}_i &=   \sigma({\bW}^{(\mathbf{i})} \tilde{\mathbf{z}}_i + \bU^{(\mathbf{i})} \tilde{\mathbf{h}}_i + \bb^{(\mathbf{i})}),\\
\mathbf{o}_i &=   \sigma({\bW}^{(\mathbf{o})} \tilde{\mathbf{z}}_i + \bU^{(\mathbf{o})} \tilde{\mathbf{h}}_i + \bb^{(\mathbf{o})}),\\
\mathbf{f}_{i,l} &=   \sigma({\bW}^{(\mathbf{f})} \bz_{i,l} + \bU^{(\mathbf{f})} {\mathbf{h}}_{i-l} + \bb^{(\mathbf{f})}),\\
\tilde{\mathbf{c}}_i &=   \phi({\bW}^{(\tilde{\mathbf{c}})} \tilde{\mathbf{z}}_i + \bU^{(\tilde{\mathbf{c}})} \tilde{\mathbf{h}}_i + \bb^{(\tilde{\mathbf{c}})}),\\
\mathbf{c}_i    &= {\tilde{\mathbf{c}}}_i \odot \mathbf{i}_i + \sum_{l\in L_i}\mathbf{c}_{i - l} \odot \mathbf{f}_{i,l}, \\
{\mathbf{h}}_i &= \mathbf{o}_i \odot \phi( \mathbf{c}_i ).
\end{align}

\subsection{Model-II: Weight Independent DAG-LSTM} \label{sec:model-II}
Since word length is a crucial information, we further take the word length into account. Specifically, the weight independent DAG-LSTMs (WI-DAG-LSTMs) could be expressed as:
\begin{align}
\mathbf{i}_i &=   \sigma(\sum_{l\in L_i}{\bW}^{(\mathbf{i})}_l \bz_{i,l} + \bU^{(\mathbf{i})}_l {\mathbf{h}}_{i-l} + \bb^{(\mathbf{i})}),\\
\mathbf{o}_i &=   \sigma(\sum_{l\in L_i}{\bW}^{(\mathbf{o})}_l \bz_{i,l} + \bU^{(\mathbf{o})}_l {\mathbf{h}}_{i-l} + \bb^{(\mathbf{o})}),\\
\mathbf{f}_{i,l} &=   \sigma({\bW}^{(\mathbf{f})}_l \bz_{i,l} + \bU^{(\mathbf{f})}_l {\mathbf{h}}_{i-l} + \bb^{(\mathbf{f})}),\\
\tilde{\mathbf{c}}_i &=   \phi(\sum_{l\in L_i}{\bW}^{(\tilde{\mathbf{c}})}_l \bz_{i,l} + \bU^{(\tilde{\mathbf{c}})}_l {\mathbf{h}}_{i-l} + \bb^{(\tilde{\mathbf{c}})}),\\
\mathbf{c}_i    &= {\tilde{\mathbf{c}}}_i \odot \mathbf{i}_i + \sum_{l\in L_i}\mathbf{c}_{i - l} \odot \mathbf{f}_{i,l}, \\
{\mathbf{h}}_i &= \mathbf{o}_i \odot \phi( \mathbf{c}_i ).
\end{align}
Notably, we will map all $\bW_l$ and $\bU_l$ to $\bW_{l_{\text{max}}}$ and $\bU_{l_{\text{max}}}$ respectively when $l > l_{\text{max}}$, which could alleviate the problem of data sparsity. Here, $l_{\text{max}}$ is a hyper-parameter.

\section{Training Strategy}
\subsection{Max-Margin criterion}
Given a train set $\mathcal{D} = \{(X_m,Y^*_m)\}_{m=1}^{M}$,  the regularized objective function $J(\Theta)$ could be expressed as:
\begin{equation}
  J(\Theta) = \frac{1}{|\mathcal{D}|} \sum_{(X_m,Y_m^*)\in \mathcal{D}} l_m(\Theta) + \frac{\lambda}{2}\|\Theta\|_2^2,
\end{equation}
where $X_m = \{x_1, \cdots, x_n\}$ and $Y^*_m = \{y_1^*, \dots, y_n^*\}$ are training character sequence and corresponding ground truth labels respectively. The loss function of each training example $l_m(\Theta)$ could be formalized as:
\begin{gather}
l_m(\Theta) = \max (0, s(X_m, \hat{Y}_m; \Theta)+\notag\\
    \Delta(Y_m^*, \hat{Y}_m)-s(X_m, Y_m; \Theta)),
\end{gather}
where $\hat{Y}_m = \{\hat{y}_1, \dots, \hat{y}_n\}$ is the the predicted labels of $m$-th training case, and is derived by a Viterbi algorithm:
\begin{equation}
\hat{Y}_m = \argmax_{Y \in \mathcal{T}^n} s (X, Y;\Theta), \label{eq:argmax}
\end{equation}
where $\mathcal{T}=\{B, M, E, S\}$.
The structured margin loss $\Delta(Y_m^*, \hat{Y}_m)$ is defined as:
\begin{equation}
    \Delta(Y_m^*, \hat{Y}_m) = \sum_{i=1}^n \eta \textbf{1} \{y_i \neq \hat{y}_i\},
\end{equation}
where $n$ is the length of $m$-th training example and $\eta$ is a discount parameter.

In this paper, we adopt AdaGrad \cite{duchi2011adaptive} with minibatchs to minimize the objective $J(\Theta)$.

\subsection{Dropout \& IV Word Dropout}
Dropout is one of prevalent methods to avoid overfitting in neural networks \cite{srivastava2014dropout} during the training phase. In this paper, we not only employ the conventional dropout strategy, but also propose a IV (in-vocabulary) word dropout strategy for DAG-LSTM. The conventional dropout strategy randomly drops out some neurons in the network with a fixed probability $p$ (dropout rate). The incoming and outgoing connections of those dropped neurons will be temporarily removed. Unlike conventional dropout, the proposed IV word dropout strategy randomly drops out the input information of some edges (NOT drop edges) in the DAG with a fixed probability $p_{\text{IV}}$ (IV word dropout rate). Specifically, the words associated with the chosen edges will be temporarily mapped to ``<OOV>'' (a special symbol indicating the unknown word). For the IV word dropout strategy, we only focus on words with multiple characters, and we reserve all single characters during training.

There are two main advantages of IV word dropout strategy. (1) Since ``<OOV>'' words will not be constructed in the DAG, the original DAG-LSTM could not deal with the OOV word in test phase. By using IV word dropout strategy, DAG-LSTM could easily exploit the OOV words in an external vocabulary to boost the performance. Specifically, the OOV words will be mapped to the ``<OOV>'' symbol when testing. (2) The IV word dropout strategy will alleviate the overfitting problem as well.

\paragraph{A Special Case} \label{para:special}
Notably, the proposed IV word dropout strategy has a very special case when the IV word dropout rate $p_{\text{IV}} = 100\%$ (drops all). In this case, only the embeddings of the single characters (as well as some special symbols, like ``<OOV>'', ``<BOS>'', etc.) will be reserved and  optimized in the training phase. And in the test phase, all the words with multiple characters will be mapped to ``<OOV>''. However, the DAGs for the train set, development set and test set will not be altered.
\begin{table} \setlength{\tabcolsep}{3pt}\small
\centering
\begin{tabular}{|c|c|}
  \hline
  Character embedding size &$d_e = 100$\\
  Initial learning rate &$\alpha = 0.2$\\
  Loss weight coefficient &$\lambda = 0.05$\\
    Margin loss discount &$\eta = 0.2$\\
  LSTM dimensionality & $d_h = 150$\\
  Conventional dropout rate &$p = 20\%$\\
  IV word dropout rate &$p_{\text{IV}} = 50\%$\\
  Batch size&$128$\\
  \hline
\end{tabular}
\caption{Configurations of Hyper-parameters.}\label{tab:paramSet}
\end{table}

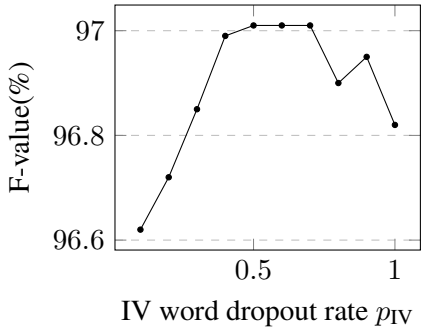
\begin{figure}[t]
  \centering
  \pgfplotsset{width=0.35\textwidth}
  \begin{tikzpicture}
    \begin{axis}[
    xlabel={IV word dropout rate $p_{\text{IV}}$},
    ylabel={F-value(\%)},
    mark size=1.0pt,
    ymajorgrids=true,
    grid style=dashed,
    legend pos= south east,
    legend style={font=\tiny,line width=.5pt,mark size=.5pt,
            at={(0.99,0.01)},
            legend columns=1,
            /tikz/every even column/.append style={column sep=0.5em}},
    ]
    \addplot [black,mark=*] table [x index=0, y index=1] {dev.txt};
    \end{axis}
\end{tikzpicture}

\caption{Performance of using different IV word dropout rate $p_{\text{IV}}$ on DAG-LSTM (Model-II) on the development set of MSRA dataset. $p_{\text{IV}}$ denotes how much drops.}\label{fig:dev}
\end{figure}
\subsection{Hyper-Parameter Configuration}
Table \ref{tab:paramSet} gives the hyper-parameter settings. Specifically, we employ conventional dropout strategy after embedding layer with dropout rate $p = 20\%$ (keeping 80\% inputs). According to the results on Figure \ref{fig:dev}, the IV word dropout rate is better set to $p_{\text{IV}} = 50\%$ (with 50\% input kept).

For initialization, all parameters is drawn from a uniform distribution $(-0.05, 0.05)$. Following previous works \cite{chen2015long,pei2014maxmargin}, all experiments including baseline results use the pre-tarined embeddings\footnote{The embeddings are pre-trained on Chinese Wikipedia corpus with word2vec toolkit \cite{mikolov2013efficient}} for initialization.

\section{Experiments}
\begin{table}\small
\centering
\begin{tabular}{|c|c|r|r|r|}
 \hline
 \multicolumn{2}{|c|}{} & \multicolumn{1}{c|}{MSRA} & \multicolumn{1}{c|}{AS} & \multicolumn{1}{c|}{CTB}\\
 \hline
  \multirow{3}{*}{Train set} & \#sent &86.9K&709.0K&23.4K\\
  &\#token&2.4M&5.4M&0.6M\\
  &\#char&4.1M&8.4M&1.1M\\
  \hline
  \multirow{3}{*}{Test set} & \#sent &4.0K&14.4K&2.1K\\
  &\#token&0.1M&0.1M&0.1M\\
  &\#char&0.2M&0.2M&0.1M\\
  \hline
\multicolumn{2}{|c|}{OOV Rate}&2.6\%  &4.3\%  &5.6\%\\
\hline
 \end{tabular}
 \caption{Details of three datasets. \#sent, \#token and \#char indicate the numbers of sentences, tokens and characters respectively. OOV Rate is the out-of-vocabulary rate, indicating how much percentage of words are there only appearing in the test set.}\label{tab:info_datasets}
\end{table}
\subsection{Datasets}
We use three prevalent datasets, MSRA, CTB and AS, to evaluate our model. The MSRA and AS are provided by SIGHAN2005 \cite{emerson2005second}, and CTB is from SIGHAN2008 \cite{moe2008fourth}. The details of the three datasets are shown in Table \ref{tab:info_datasets}. We use 10\% data of shuffled train set as development set for all datasets.

\subsection{Overall Results}
\begin{table*}[t]\setlength{\tabcolsep}{3pt}\small %
\centering
\begin{tabular}{|p{54pt}|c|*{12}{c|}}
\hline
 \multirow{2}*{Models} &\multicolumn{4}{c|}{MSRA}&\multicolumn{4}{c|}{AS}&\multicolumn{4}{c|}{CTB}\\
  \cline{2-13}
  &P    &R  &F  &OOV    &P    &R  &F  &OOV    &P    &R  &F  &OOV    \\
\hline
\hline
\multicolumn{13}{|l|}{Baselines}\\
\hline
Uni-gram&	94.19 &	94.08 &	94.14 &	65.33 &	92.13 &	92.79 &	92.46 &	69.57 &	93.68 &	93.94 &	93.81 &	75.15\\
Bi-gram&	95.59 &	95.82 &	95.71 &	65.74 &	93.64 &	94.77 &	94.20 &	70.07 &	95.17 &	95.26 &	95.22 &	75.62\\
\hline
\multicolumn{13}{|l|}{DAG-LSTM}\\
\hline
Model-I&	95.98 &	96.18 &	96.08 &	65.15 &	94.42 &	95.47 &	94.94 &	68.38 &	95.53 &	95.71 &	95.62 &	77.09\\
Model-II&	96.01 &	96.13 &	96.07 &	67.65 &	94.54 &	95.26 &	94.90 &	70.70 &	\textbf{95.56} &	95.60 &	95.58 &	76.49\\
\hline
\multicolumn{13}{|l|}{DAG-LSTM with IV word dropout strategy (IV word dropout rate $p_{\text{IV}} = 50\%$)}\\
\hline
Model-I&	95.96 &	96.35 &	96.15 &	64.51 &	94.31 &	\textbf{95.58} &	94.94 &	67.38 &	95.52 &	95.51 &	95.52 &	77.16\\
Model-II&	\textbf{96.40} &	\textbf{96.53} &	\textbf{96.47} &	\textbf{68.54} &	\textbf{94.86} &	95.49 &	\textbf{95.17} &	\textbf{71.33} &	95.55 &	\textbf{95.73} &	\textbf{95.64} &	\textbf{76.96}\\
\hline
\hline
\multicolumn{13}{|c|}{The special case: DAG-LSTM with IV word dropout rate $p_{\text{IV}} = 100\%$ (drops all)}\\
\hline
Model-I&	94.59 &	94.66 &	94.62 &	\textbf{62.64} &	\textbf{94.41} &	\textbf{95.05} &	\textbf{94.73} &	\textbf{73.09} &	95.22 &	95.16 &	95.19 &	\textbf{78.42}\\
Model-II&	\textbf{94.86} &	\textbf{94.94} &	\textbf{94.90} &	62.27 &	93.87 &	94.80 &	94.33 &	72.52 &	\textbf{95.30} &	\textbf{95.49} &	\textbf{95.39} &	77.59\\
\hline
\hline
\multicolumn{13}{|c|}{Using the vocabulary of the test set $\mathcal{V}_{\text{test}}$}\\
\hline
Model-I&	96.88 &	96.83 &	96.86 &	74.26 &	96.36 &	\textbf{96.67} &	96.52 &	82.17 &	96.45 &	96.13 &	96.29 &	85.95 \\
Model-II&	\textbf{97.35} &	\textbf{96.88} &	\textbf{97.11} &	\textbf{82.10} &	\textbf{96.77} &	96.66 &	\textbf{96.71} &	\textbf{85.80} &	\textbf{96.86} &	\textbf{96.44} &	\textbf{96.65} &	\textbf{88.10} \\
\hline
\end{tabular}

\caption{Results of the proposed models on test sets of three datasets. P, R, F and OOV indicate precision, recall, F value and out-of-vocabulary recall rate respectively. The maximum F values in each block is highlighted for each dataset.}\label{tab:res}
\end{table*}
Table \ref{tab:res} gives the overall results of our model on test sets of three CWS datasets, which consists of three main blocks.

\paragraph{The First Block} The first main block contains three sub-blocks: two baseline models, DAG-LSTM and DAG-LSTM with IV word dropout strategy.

(1) The first sub-block contains two baselines: Bi-LSTM models using uni-gram and bi-gram respectively (mentioned in Section \ref{sec:unigram} and Section \ref{sec:bigram}). As previous work \cite{pei2014maxmargin,chen2015long} reports, the performance can be significantly boosted by using bi-gram features, since the bi-gram model additionally exploit the information of words with two characters. As shown in Table \ref{tab:res}, the model with bi-gram feature boosts +1.57\%, +1.74\% and +1.41\% on F value on MSRA, AS and CTB respectively.

(2) The second sub-block gives the results of the proposed DAG-LSTM model, where Model-I indicates the weight sharing DAG-LSTM (WS-DAG-LSTM) (Section \ref{sec:model-I}) and Model-II indicates the weight independent DAG-LSTM (WI-DAG-LSTM) (Section \ref{sec:model-II}). By using DAG-LSTM, we obtain significant improvement on performance. As shown in Table \ref{tab:res}, the DAG-LSTM model obtains 96.08\%, 95.47\% and 95.62\% on F value on MSRA, AS and CTB respectively. Compared to bi-gram model, the performance boosts +0.37\%, +0.74\% and +0.40\% respectively, since the proposed DAG-LSTM model benefit from both the word level (not only the words with two characters) and the character level information. We could also observe that the performance of Model-I and Model-II is comparable. Strictly speaking, the performance of Model-I is slight better than Model-II. However, the number of trainable parameters of Model-I is much less than Model-II, since the Model-I shares the weight matrix over various input. It might be caused by the problem of overfitting, so that Model-II performances poorly.

(3) The third sub-block shows the effectiveness of the proposed IV word dropout strategy on the  proposed DAG-LSTM model. As we can see, the performance further boosts. Besides, by using the IV word dropout strategy, Model-II boost significantly, and outperforms Model-I, whereas the effects of the IV word dropout strategy on Model-I is modest. It shows that the IV word dropout strategy alleviates the problem of overfitting of Model-II.

\paragraph{The Second Block} The second main block shows the results of the special case where IV word dropout rate $p_{\text{IV}} = 100\%$ (mentioned in Section \ref{para:special}).
According to the results of the special case, we could observe that the performance on all datasets significantly outperforms the baseline LSTM model with uni-gram feature (from 94.14\% to 94.90\%, from 92.46\% to 94.73\% and from 93.81\% to 95.39\% on F value on MSRA, AS and CTB respectively). Moreover, the performance outperforms the bi-gram model on AS and CTB datasets as well. It shows that the proposed DAG-LSTM could well model the word level information by only using the information ``whether there is a word'' instead of ``what the word is there''. In an other word, DAG-LSTM could also perform well when only character embeddings (as well as some special symbols, like ``<OOV>'', ``<BOS>'', etc.) are available (NO embeddings of words with multiple characters).

\paragraph{The Third Block} The experimental settings of the third main block are the same with the third sub-block of the first main block (using IV word dropout strategy and  $p_{\text{IV}} = 50\%$), but additionally exploit the vocabulary of the test set $\mathcal{V}_{\text{test}}$. As shown in Table \ref{tab:info_datasets}, the OOV rates of three datasets are all very small, which means that the overlap of $\mathcal{V}_{\text{test}}$ and $\mathcal{V}_{\text{train}}$ is significant, and very few words only occur in the test set. However, the performance is significantly boosted by only introducing 2.6\%, 4.3\% and 5.6\% more words in vocabulary on MSRA, AS and CTB datasets respectively. Compared with the results of the third sub-block in the first main block, the performance boosts +0.64\%, +1.54\% and +1.01\% on F value on MSRA, AS and CTB respectively, and the OOV recall rate boosts +13.56\%, +14.47\% and +11.14\% respectively. Notably, the additionally imported words have no trained embeddings and will be mapped to ``<OOV>'' symbol for testing. It shows that the proposed DAG-LSTM model could perfectly exploit the external vocabulary without re-training, whereas it is non-trivial for previous models to exploit an external vocabulary. Concretely, the generalization ability of most of previous models are modest. Since the out-of-vocabulary words do not appear in the train set, they cannot exploit external vocabularies even if they re-train on the train set. Therefore, when we are going to use a trained segmenter to segment some corpus with large mount of out-of-vocabulary words, such as patent and medical documents, the proposed DAG-LSTM model could obtain a great boost by incorporating an external dictionary of professional terms.

\subsection{Effects of Vocabulary}
\begin{table}\setlength{\tabcolsep}{3pt}\small %
\centering
\begin{tabular}{|c|*{4}{c|}}
\hline
 \multirow{2}*{Models} &\multicolumn{4}{c|}{MSRA}\\
  \cline{2-5}
  &P    &R  &F  &OOV\\
\hline
\multicolumn{5}{|l|}{Baselines}\\
\hline
Uni-gram&	94.19 &	94.08 &	94.14 &	65.33 \\
Bi-gram&	95.59 &	95.82 &	95.71 &	65.74 \\
\hline
\multicolumn{5}{|l|}{DAG-LSTM with IV word dropout strategy}\\
\hline
Max word length = 1&	94.19 &	94.08 &	94.14 &	65.33 \\
Max word length = 2&	95.68 &	95.81 &	95.74 &	65.19 \\
Max word length = 3&	95.78 &	96.01 &	95.89 &	65.12 \\
Max word length = 4&	96.21 &	96.31 &	96.26 &	67.34 \\
All words&	\textbf{96.40} &	\textbf{96.53} &	\textbf{96.47} &	\textbf{68.54}\\
\hline
\end{tabular}

\caption{Effects of vocabulary of DAG-LSTM (Model-II with IV word dropout strategy) on the test set of MSRA dataset. The maximum values are highlighted for each colunm.}\label{tab:vocab}
\end{table}

We also investigate the effects of vocabulary to the proposed DAG-LSTM model. The vocabulary is extracted from gold segmentation of the train set for each dataset, which is given as a part of corpus. Table \ref{tab:vocab} gives the results of DAG-LSTM (Model-II) with IV word dropout rate $p_{\text{IV}} = 50\%$ (the same configuration with the third sub-block of the third main block in Table \ref{tab:res} except the vocabulary) on the test set of MSRA dataset. We experiment the cases when we only keep partial words of the given vocabulary, providing words whose lengths are not greater than a given ``Max word length''. As shown in Table \ref{tab:vocab}, we tries to set the max word length from 1 to 4, and we also report the performance of DAG-LSTM using the whole vocabulary (whose results is the same with Model-II in the third sub-block of the first main block in Table \ref{tab:res}). As we can see in Table \ref{tab:vocab}, the performance boosts gradually when we exploit much more words of the given vocabulary, and the model performances best when the whole vocabulary is employed. Specially, the results of the case (Max word length = 1) is the same with unigram model, since they are really the same model and the unigram model could be viewed as a special case of the proposed DAG-LSTM model. Moreover, we could observe that the case (Max word length = 2) outperforms the bigram model, since we additionally incorporate the long term information through the DAG-LSTM.

\subsection{Case Study}
Table \ref{tab:case} gives two cases. The first one is from the 1521-th example in the test set of  CTB and the secode one is from the 2626-th example in the test set of MSRA. The baselines in two cases all denote the Bi-LSTM model with bigram feature.

(1) In the first case, our model is DAG-LSTM with IV word dropout strategy. Since ``adulthood'' is an in-vocabulary (IV) word, our model could segment it corrected by utilizing the word information via DAG-LSTM.

(2) In the second case, we additionally use the vocabulary of the test set $\mathcal{V}_{\text{test}}$ to build DAG. Here, `` subsidy'' is an IV word, and `` subsidy rate'' is an out-of-vocabulary (OOV) word. It shows that the proposed DAG-LSTM model could benefit from a given vocabulary to improve the performance.

\begin{table}[t]\setlength{\tabcolsep}{3pt}\small
%
\centering
  \includegraphics[width=0.45\textwidth]{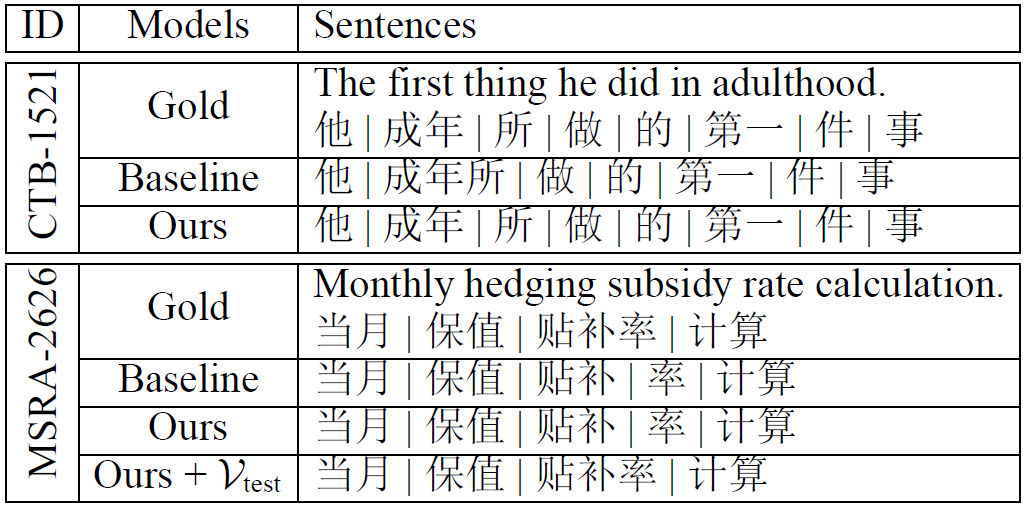}
\caption{Case Study.}\label{tab:case}
\end{table}

\section{Related Work}

Chinese word segmentation has been studied with considerable efforts in the NLP community. Specific to the word-based CWS, some pioneering work adopts Semi-Markov CRF \cite{Andrew:2006} or transition-based model \cite{zhang2007chinese}. Recently, several neural models are proposed to utilize the word-level information.

\citet{cai2016neural} formalize word segmentation as a direct structured learning procedure. Specifically, they employ a gated combination neural network over characters and a LSTM over words to calculate the score for each candidate segmentation.

\citet{zhang2016transition} propose a neural model to utilize the word-level information under the transition-based framework \cite{zhang2007chinese}. Their model exploits not only character embeddings as previous work does, but also word embeddings pre-trained from large scale corpus.

\citet{liu2016exploring} follow the work of \cite{Andrew:2006} and use a semi-CRF taking segment-level embeddings as input.

All these word-based CWS are not character based sequence labeling. Their inference are inexact with the beam-search, and they are constrained by the word length.

\section{Conclusion}\label{sec:conclusion}

In this paper, we propose a character-based model, DAG-LSTM, for neural word segmentation to incorporate the word-level information. Our method can further boost the performance of CWS by using an external vocabulary, which is essential in practice. Experiments show that our proposed model outperforms the baseline methods on three popular benchmarks.


\bibliography{emnlp2017}
\bibliographystyle{emnlp_natbib}

\end{document}